\documentclass{article}

\usepackage{PRIMEarxiv}

\usepackage[utf8]{inputenc} 
\usepackage[T1]{fontenc}    
\usepackage{hyperref}       
\usepackage{url}            
\usepackage{booktabs}       
\usepackage{amsfonts}       
\usepackage{nicefrac}       
\usepackage{microtype}      
\usepackage{lipsum}
\usepackage{fancyhdr}       
\usepackage{graphicx}       
\graphicspath{{media/}}
\usepackage{amsmath}
\usepackage[table,xcdraw]{xcolor}

\title{Which Backbone to Use: A Resource-efficient Domain Specific Comparison for Computer Vision
}

\author{
  Pranav Jeevan P \hspace{2cm} Amit Sethi \\
  Department of Electrical Engineering \\
  Indian Institute of Technology Bombay \\
  Mumbai, India  \\
  \texttt{\{194070025, asethi\}@iitb.ac.in} \\
}

\begin{document}
\maketitle

\begin{abstract}
In contemporary computer vision applications, particularly image classification, architectural backbones pre-trained on large datasets like ImageNet are commonly employed as feature extractors. Despite the widespread use of these pre-trained convolutional neural networks (CNNs), there remains a gap in understanding the performance of various resource-efficient backbones across diverse domains and dataset sizes. Our study systematically evaluates multiple lightweight, pre-trained CNN backbones under consistent training settings across a variety of datasets, including natural images, medical images, galaxy images, and remote sensing images. This comprehensive analysis aims to aid machine learning practitioners in selecting the most suitable backbone for their specific problem, especially in scenarios involving small datasets where fine-tuning a pre-trained network is crucial. Even though attention-based architectures are gaining popularity, we observed that they tend to perform poorly under low data fine-tuning tasks compared to CNNs. We also observed that some CNN architectures such as ConvNeXt, RegNet and EfficientNet performs well compared to others on a diverse set of domains consistently. Our findings provide actionable insights into the performance trade-offs and effectiveness of different backbones, facilitating informed decision-making in model selection for a broad spectrum of computer vision domains. Our code is available here: \url{https://github.com/pranavphoenix/Backbones}
\end{abstract}

\keywords{Image Classification \and Backbone \and Fine tuning \and Domains}

\begin{table}[b]
\centering
\resizebox{\textwidth}{!}{%
\begin{tabular}{@{}llllllll@{}}
\toprule
\multicolumn{8}{c}{\textbf{Domains}} \\ \midrule
\textbf{Ranking} & \multicolumn{1}{c}{\textbf{Natural}} & \multicolumn{1}{c}{\textbf{Texture}} & \multicolumn{1}{c}{\textbf{Remote Sensing}} & \multicolumn{1}{c}{\textbf{Plant}} & \multicolumn{1}{c}{\textbf{Astronomy}} & \multicolumn{1}{c}{\textbf{Medical}} & \multicolumn{1}{c}{\textbf{Overall}} \\ \midrule
\rowcolor[HTML]{FFFFFF} 
\textbf{Best} & \textbf{ConvNeXt-Tiny} & \textbf{ConvNeXt-Tiny} & \textbf{ResNeXt-50 $32\times4$d} & \textbf{RegNetY-3.2GF} & \textbf{WaveMix} & \textbf{WaveMix} & \textbf{ConvNeXt-Tiny} \\
\rowcolor[HTML]{FFFFFF} 
Better & EfficientNetV2-S & ResNeXt-50 $32\times4$d & EfficientNetV2-S & ConvNeXt-Tiny & ConvNeXt-Tiny & EfficientNetV2-S & RegNetY-3.2GF \\
\rowcolor[HTML]{FFFFFF} 
Good & RegNetY-3.2GF & RegNetY-3.2GF & ConvNeXt-Tiny & ShuffleNetV2 $2.0\times$ & DenseNet-161 & RegNetY-3.2GF & EfficientNetV2-S \\ \bottomrule
\end{tabular}%
}
\vspace{2mm}
\caption{A summary of our observations showing the top 3 backbones for fine-tuning on multiple domains for image classification}
\label{tab:summary}
\end{table}

\section{Introduction}

In most computer vision applications, particularly in image classification, practitioners commonly use a backbone feature extraction network paired with a classification head. This approach involves using the backbone to output features from the input image, which are then fed into a task-specific head for further computations. A prevalent practice in the field is to leverage backbones trained on large datasets, such as ImageNet-1k~\cite{5206848} and then fine-tune them on smaller datasets. This strategy is particularly useful due to the scarcity of large, domain-specific datasets, such as those in the medical applications.

Practitioners often resort to using off-the-shelf models from libraries like Torchvision~\cite{torchvision2016}, which offers a plethora of backbones with pre-trained ImageNet weights. These pre-trained backbone architectures, when fine-tuned on smaller, domain-specific datasets, consistently outperform models trained from scratch. This method not only enhances performance but also reduces the computational resources and time required for training, making it a dominant procedure in the field of computer vision.

However, it is important to note that while Torchvision lists the performance metrics (top-1 accuracy) of various backbone architectures on the ImageNet dataset, these metrics might not necessarily translate to similar performance rankings in fine-tuning datasets. For instance, a backbone with higher ImageNet performance might not guarantee superior performance on a particular fine-tuning dataset. Conversely, a backbone with weaker ImageNet performance might outperform others when fine-tuned on a dataset which belongs to a different domain. This possible discrepancy highlights the need for careful selection and evaluation of backbone models in various fine-tuning datasets belonging to different domains, underscoring the complexity and variability of transfer learning in computer vision.

Moreover, one of the major challenges for practitioners is the limitation in terms of resources. These limitations include GPU availability, training and inference time, and even model size. To address these challenges, our experiments focus on comparing only the light-weight (model size less than 100 MB) backbone architectures available on Torchvision, which are designed to be resource-efficient while maintaining high inference speed.

The lack of benchmarks that fit these backbones across multiple domains presents a significant challenge for practitioners in deciding which backbone to use for specific datasets with limited data. Additionally, the size of the fine-tuning dataset can also impact the performance of these backbone architectures. Our study also aims to investigate whether the dataset size influences the selection of backbone models and to provide insights on choosing the most suitable backbone based on the specific domain and the amount of available fine-tuning data.

We also analyse the architectures which perform well in each of the domain-specific datasets and try to understand their properties. This will help the research community to realise their fundamental limitations and design better architectures. 

\section{Related Works}
\label{sec:related}

Most of the research on architectures uses ImageNet benchmark to compare image classification performance of backbones. Pytorch Image Models~\cite{rw2019timm} library benchmarks ImageNet classification performance across different backbone architectures. Visual Task Adaptation Benchmark~\cite{zhai2020largescale} evaluated the performance of computer vision models on 19 tasks that span different domains. Brigato et al.~\cite{brigato2021tune} designed a benchmark for data-efficient image classification spanning various domains and showed that tuning hyper-parameters like learning rate, weight decay, and batch size results in a competitive baseline outperforming most specialized methods. Taher et al.~\cite{taher2021systematic} benchmarked various self-supervised pre-training methods, 14 pre-trained ImageNet models on 7 diverse medical tasks. Battle of backbones~\cite{goldblum2023battle} benchmarks few pre-trained ImageNet backbones, including those pre-trained by self-supervised learning and Stable Diffusion, across a diverse set of computer vision tasks ranging from image classification, Out-of-distribution generalisation, image retrieval and object detection.   
There is a lack of bench-marking on transfer learning performance of popular lightweight backbones across multiple domain-specific datasets.

\section{Backbone Architectures}
\label{sec:architecture}

Since most of the practitioners use out of the shelf ImageNet pre-trained models from the popular Torchvision library, we selected models from this library for our experiments. We used certain criteria to filter models keeping in mind our resource-efficiency constraint. So, we decided to choose models with number of parameters less than 30 million, which amounts to a storage space of around 100 MB. WaveMix~\cite{jeevan2024wavemix} model was added to our list of backbones because WaveMix was shown to provide state-of-the-art (SOTA) performance in multiple image classification datasets belonging to many diverse domains such as EMNIST~\cite{cohen2017emnist} and galaxy morphology datasets. 

Torchvision has 20 backbone families with models of multiple sizes within each family making the total available models 115. After we apply our constraint of lightweight models (less than 30 M parameters), we have 53 models belonging to 14 backbone families. We decided to take the best model (highest ImageNet top-1 accuracy) from each of these families for our experiments. We also discarded models whose ImageNet top-1 accuracy was less than 75\%. 

The final list of selected 11 backbones for our experiments are shown in table~\ref{tab:imagenet}. We added both Swin-Tiny~\cite{liu2021swin} and SwinV2-Tiny~\cite{liu2022swin} in our list because there was an under-representation of attention-based transformer architectures in our list and the only other attention based model in Torchvision was vision-transformer (ViT) whose smallest model has 86 million parameters. We wanted to also understand how attention-based models perform in resource-efficient fine-tuning under low data regime since there has been a recent trend amoung computer vision practitioners to use transformers for all tasks.

\begin{table}[]
\centering
\begin{tabular}{@{}lrr@{}}
\toprule
Architectures & \begin{tabular}[c]{@{}r@{}}\# Params\\ (M)\end{tabular} & \begin{tabular}[c]{@{}r@{}}ImageNet-1k\\ Top-1 Accuracy\\ (\%)\end{tabular} \\ \midrule
ResNet-50~\cite{he2015deep} & 25.6 & 76.13 \\
WaveMix~\cite{jeevan2024wavemix} & 27.9 & 75.32 \\
ConvNeXt-Tiny~\cite{liu2022convnet} & 28.6 & 82.52 \\
Swin-Tiny~\cite{liu2021swin} & 28.3 & 81.47 \\
SwinV2-Tiny~\cite{liu2022swin} & 28.4 & 82.07 \\
EfficientNetV2-S~\cite{tan2021efficientnetv2} & 21.5 & 84.23 \\
DenseNet-161~\cite{huang2018densely} & 28.7 & 77.14 \\
MobileNetV3-Large~\cite{howard2019searching} & 5.5 & 75.27 \\
RegNetY-3.2GF~\cite{radosavovic2020designing} & 19.4 & 81.98 \\
ResNeXt-50~\cite{xie2017aggregated} $32\times4$d & 25.0 & 81.20 \\
ShuffleNetV2~\cite{ma2018shufflenet} $2.0\times$ & 7.4 & 76.23 \\ \bottomrule
\end{tabular}
\vspace{5mm}
\caption{List of popular computer vision backbones we used for our experiments along with the model size and ImageNet accuracy for which model weights are available. All model weights are taken from Torchvision library except for WaveMix which was taken from GitHub. }
\label{tab:imagenet}
\end{table}

Below are the architectures which was selected for our experiments.

\textbf{ResNet~\cite{he2015deep}}: ResNets are the most popular and successful backbone architectures currently in use today since its arrival almost a decade ago. ResNet uses residual connections to allow for the training of very deep convolutional neural networks (CNN) by mitigating the vanishing gradient problem. We use ResNet-50 for our experiments.

\textbf{WaveMix~\cite{jeevan2024wavemix}}: WaveMix is a token-mixing architecture that uses 2-dimensional discrete wavelet transform for spacial token-mixing and has been shown to provide SOTA performance in multiple image classification datasets. We use WaveMix-192/16 (level 3) for our experiments.

\textbf{ConvNeXt~\cite{liu2022convnet}}: ConvNeXt is a recent CNN architecture designed to improve upon traditional CNNs by incorporating elements from transformer models, resulting in enhanced performance and scalability. It uses depth-wise convolutions, inverted bottleneck blocks and large kernels. We use ConvNeXt-Tiny in our experiments.

\textbf{Swin Transformer~\cite{liu2021swin}}: Swin transformer was an improvement over conventional ViT which overcame the massive data requirements for training. It incorporated efficiency using hierarchical representations, limiting the attention window and merging them stage by stage. We use Swin-Tiny and SwinV2-Tiny~\cite{liu2022swin} for our experiments. 

\textbf{EfficientNet~\cite{tan2021efficientnetv2}}: EfficientNet is a family of CNNS that optimize both model size and speed by utilizing a compound scaling method that uniformly scales network depth, width, and resolution. It incorporates advanced techniques such as progressive learning and a mix of regular and mobile convolutions. We use EfficientNetV2-S in our experiments.

\textbf{Densenet~\cite{huang2018densely}}: DenseNet is a CNN architecture that connects each layer to every other layer in a feed-forward fashion, promoting feature reuse and reduction in number of parameters. This dense connectivity pattern helps alleviate the vanishing gradient problem and leads to improved training efficiency and accuracy. We use Densenet-161 in our experiments.

\textbf{MobileNet~\cite{howard2019searching}}: MobileNetV3 is a CNN architecture designed for on-device and resource-constrained environments, which combines lightweight depth-wise separable convolutions with squeeze-and-excitation modules. It is a highly efficient model with improved accuracy and reduced computational complexity. We use MobileNetV3-Large in our experiments.

\textbf{RegNet~\cite{radosavovic2020designing}}: RegNet is a family of CNN architectures that utilize a regular design space to systematically generate a diverse range of models, optimizing for both efficiency and performance. It focuses on simple, scalable structures with uniform depth, width, and group convolution patterns, incorporating features like bottleneck blocks. We use RegNetY-3.2GF for our experiments.

\textbf{ResNeXt~\cite{xie2017aggregated}}: ResNeXt is a CNN architecture that extends the ResNet model by introducing a cardinality dimension, using grouped convolutions to aggregate multiple transformations, which improves performance and efficiency. We use ResNeXt-50 $32\times4$d for our experiments.

\textbf{ShuffleNet~\cite{ma2018shufflenet}}: ShuffleNet is a lightweight CNN architecture designed for efficient computation on mobile devices. It uses channel shuffling and point-wise group convolution to optimize speed and accuracy. We use ShuffleNetV2 $2.0\times$ for our experiments.

\section{Datasets}

Since we wanted to check the performance of various backbones on data efficient fine-tuning, we decided to perform our experiments on datasets with utmost 100,000 training images. We choose 20 publicly available datasets belonging to 7 different domains such as natural, textures, remote sensing, plants, astronomy, surface defect and medical images with number of training images ranging from 1000 to 100,000 and number of classes ranging from 2 to 200.  Details of datasets in each domain is given in table~\ref{tab:dataset}. 

\begin{table}[]
\centering
\begin{tabular}{@{}llrrr@{}}
\toprule
Datasets & Domain Description & \begin{tabular}[c]{@{}r@{}}\# Training \\ Images\end{tabular} & \begin{tabular}[c]{@{}r@{}}\# Testing\\ Images\end{tabular} & \# Classes \\ \midrule
CIFAR-10~\cite{Krizhevsky09learningmultiple} & Natural Images & 50,000 & 10,000 & 10 \\
CIFAR-100~\cite{Krizhevsky09learningmultiple} & Natural Images & 50,000 & 10,000 & 100 \\
Tiny ImageNet~\cite{Le2015TinyIV} & Natural Images (ImageNet subset) & 100,000 & 10,000 & 200 \\
Stanford dogs~\cite{KhoslaYaoJayadevaprakashFeiFei_FGVC2011} & Natural Images (Dog breeds) & 12,000 & 8,580 & 120 \\
Flowers-102~\cite{Nilsback08} & Natural Images (Flower species) & 2,040 & 6,149 & 102 \\
CUB-200-2011~\cite{WelinderEtal2010} & Natural Images (Bird species) & 5,994 & 5,794 & 200 \\
Stanford Cars~\cite{dehghan2017view} & Natural Images (Car models) & 8,144 & 8,041 & 196 \\
Food-101~\cite{10.1007/978-3-319-10599-4_29} & Natural Images (Food categories) & 75,750 & 25,250 & 101 \\
DTD~\cite{cimpoi14describing} & Texture Images & 1,880 & 1,880 & 47 \\
NEU Surface Defects~\cite{SONG2013858} & Surface Defect Images & 1,440  & 360 & 6 \\
UC Merced Land Use~\cite{Nilsback08uc} & Remote Sensing Images & 1,680 & 420 & 21 \\
EuroSAT~\cite{helber2019eurosat} & Remote Sensing Images & 18,900 & 8,100 & 10 \\
PlantVillage~\cite{hughes2016open} & Plant Images & 44,343 & 11,105 & 39 \\
PlantCLEF~\cite{goeau2021overview} & Plant Images & 10,455 & 1135 & 20 \\
Galaxy10 DECals~\cite{Leung_2018} & Astronomy Images (Galaxy Morphology) & 15,962 & 1,774 & 10 \\
BreakHis $40\times$~\cite{7312934} & Medical Images (Histopathology) & 1,398 & 606 & 2 \\
BreakHis $100\times$~\cite{7312934} & Medical Images (Histopathology) & 1,458 & 632 & 2 \\
BreakHis $200\times$~\cite{7312934} & Medical Images (Histopathology) & 1,411 & 611 & 2 \\
BreakHis $400\times$~\cite{7312934} & Medical Images (Histopathology) & 1,276 & 553 & 2 \\ 
RSNA Pneumonia Detection~\cite{rsna-pneumonia-detection-challenge} & Medical Images (Radiology) & 24,181 & 6046 & 2 \\  \bottomrule
\end{tabular}
\vspace{5mm}
\caption{Details of the datasets used specifying the domain details, number of images in the training and testing set and the number of classes for classification}
\label{tab:dataset}
\end{table}

\subsection{Natural}

\textbf{CIFAR-10~\cite{Krizhevsky09learningmultiple}}: CIFAR-10 dataset is a widely-used benchmark dataset for image classification, consisting of 60,000 $32\times32$ color images in 10 different classes, with 6,000 images per class divided into 50,000 training images and 10,000 test images. 

\textbf{CIFAR-100~\cite{Krizhevsky09learningmultiple}}: CIFAR-100 dataset uses the same images as CIFAR-10, but images are distributed across 100 different classes, with 600 images per class divided into 50,000 training images and 10,000 test images. It provides a more challenging classification task compared to CIFAR-10 due to the larger number of classes.

\textbf{Tiny ImageNet~\cite{Le2015TinyIV}}: Tiny ImageNet dataset is a subset of the ImageNet dataset, consisting of 200 image classes with 500 training images and 50 test images per class, each resized to $64\times64$ pixels. It is widely used for benchmarking image classification algorithms, particularly in low-resource scenarios.

\textbf{Stanford Dogs~\cite{KhoslaYaoJayadevaprakashFeiFei_FGVC2011}}: Stanford Dogs dataset is a comprehensive dataset for fine-grained image classification, containing 20,580 images of 120 different dog breeds. It is widely used for benchmarking algorithms, particularly in distinguishing between closely related categories.

\textbf{Flowers-102~\cite{Nilsback08}}: The Flowers 102 dataset is a dataset for fine-grained image classification, consisting of 8,189 images of flowers categorized into 102 different species. Each class has between 40 to 258 images, and the dataset is commonly used to benchmark algorithms in classification tasks due to its diversity and challenging nature.

\textbf{CUB-200-2011~\cite{WelinderEtal2010}}: Caltech-UCSD Birds-200-2011 is a comprehensive dataset for fine-grained image classification, consisting of 200 bird species with 11,788 annotated images. It is widely used for benchmarking algorithms in fine-grained visual recognition tasks due to its high level of granularity.

\textbf{Stanford Cars~\cite{dehghan2017view}}: The Stanford Cars dataset is a large-scale dataset for fine-grained image classification, consisting of 16,185 images of 196 classes of cars. It is widely used for evaluating and benchmarking computer vision algorithms in tasks involving fine-grained visual recognition and object detection.

\textbf{Food-101~\cite{10.1007/978-3-319-10599-4_29}}: Food-101 dataset is a large-scale dataset for food classification, containing 101,000 images of 101 different food categories, with 750 training images and 250 test images per class. It is commonly used to benchmark image recognition algorithms in the context of food and culinary applications.

\subsection{Texture}

\textbf{DTD~\cite{cimpoi14describing}}: Describable Textures Dataset (DTD) is a collection of 5,640 texture images categorized into 47 classes based on human-describable attributes. It is used to evaluate and benchmark algorithms in texture recognition and classification tasks.

\subsection{Surface Defect}

\textbf{NEU Surface Defect~\cite{SONG2013858}}: The Northeastern University (NEU) surface defect database is a benchmark dataset for surface defect detection and classification, featuring images of six types of surface defects. The dataset includes 1,800 images with a resolution of $200\times200$ pixels, divided into 1,200 training images and 600 testing images. Each defect type has an equal number of images, making it a well-balanced dataset for training and evaluating machine learning models.

\subsection{Remote Sensing}

\textbf{UC Merced Land Use~\cite{Nilsback08uc}}: UC Merced Land Use dataset is a high-resolution dataset for land use classification, containing 2,100 aerial images categorized into 21 land use classes with 100 images per class. Each image is $256\times256$ pixels, and the dataset is commonly used for evaluating and benchmarking algorithms in remote sensing and geospatial analysis tasks.

\textbf{EuroSAT~\cite{helber2019eurosat}}: The EuroSAT dataset is a benchmark dataset for land use and land cover classification, consisting of 27,000 RGB and multi-spectral images covering 10 classes, with images derived from Sentinel-2 satellite data. It is widely used for evaluating the performance of machine learning algorithms in remote sensing and geospatial analysis tasks.

\subsection{Plant}

\textbf{PlantVillage~\cite{hughes2016open}}: PlantVillage dataset is a comprehensive dataset for plant disease classification, containing over 54,000 images of healthy and diseased leaves across 39 different plant categories. It is widely used for benchmarking machine learning algorithms in agricultural and plant pathology applications.

\textbf{PlantCLEF~\cite{goeau2021overview}}: The PlantCLEF dataset is a large-scale dataset for plant identification, comprising millions of images covering thousands of plant species, including trees, flowers, fruits, and leaves. It is used for evaluating and benchmarking algorithms in botanical classification and plant biodiversity studies. We use a subset of this dataset. 

\subsection{Astronomy}

\textbf{Galaxy 10 DECals~\cite{Leung_2018}}: Galaxy 10 DECals dataset is a dataset for galaxy classification, consisting of 17,000 images of galaxies classified into 10 different morphological categories. It is used for evaluating and benchmarking machine learning algorithms in astronomy and astrophysical research.

\subsection{Medical}

\textbf{BreakHis~\cite{7312934}}: Breast Cancer Histopathology Database is a dataset specifically designed for the classification of breast cancer histopathological images. It contains 7,909 microscopic images of breast tumor tissue, divided into benign and malignant categories. It provides microscopic images of breast tumor tissue at four different magnification levels: $40\times$, $100\times$, $200\times$ and $400\times$. Each magnification level offers a different level of detail, allowing for a comprehensive analysis of histopathological features. The dataset is widely used for evaluating and bench-marking algorithms in medical image analysis and computer-aided diagnosis.

\textbf{RSNA Pneumonia Detection Challenge Dataset~\cite{rsna-pneumonia-detection-challenge}}: The RSNA Pneumonia Detection Challenge dataset, provided by the Radiological Society of North America, consists of over 30,000 chest X-ray images annotated for the presence of pneumonia. It is designed to facilitate the development of machine learning models for pneumonia detection, promoting advancements in medical imaging analysis.

\section{Experimental  Details}

Since the task is to measure fine-tuning performance of these backbones in image classification via transfer learning, a standard training protocol was used to compare the performance of all models. We do not freeze any layers of the pre-trained backbones and fine-tune the full model on the given dataset or a fraction of it, and measure the top-1 accuracy in the test set. Only the final linear layer of the backbone was modified to match the number of classes of each fine-tuning dataset.

All images were resized to $256\times256$ for our experiments, except for BreakHis fine-tuning where we resized images to $672\times448$ since reducing the resolution of histopathology images lead to poor results across models. TrivialAugment~\cite{müller2021trivialaugment} was used as data augmentation for all datasets except BreakHis, Galaxy 10 DEcals,  Augmentation was only used after checking whether using this augmentation improved the model performance or not.

Due to limited computational resources, we use early stopping where training is stopped if accuracy stops improving after 10 epochs. All experiments were done on a single 80 GB Nvidia A100 GPU. Training was done using DualOpt~\cite{jeevan2022convolutional} where we used AdamW optimizer ($\alpha = 0.001, \beta_{1} = 0.9, \beta_{2}=0.999, \epsilon = 10^{-8}$) with a weight decay of 0.01 during the initial phase and then used SGD with a learning rate of $0.001$ and momentum $= 0.9$ during the final phase.

Cross-entropy loss was used for image classification. Batch-size was chosen so that the entire GPU will be utilized while training. We used automatic mixed precision in PyTorch during training. Top-1 accuracy on the test set of best of three runs with random initialization is reported as a generalization metric based on prevailing protocols~\cite{hassani2021escaping}. 

\section{Results}

\subsection{Fine-tuning Performance}
The results of our fine-tuning experiments on datasets of all domains are shown in Table~\ref{tab:natural}, Table~\ref{tab:domain} and Table~\ref{tab: medical}. We find that ConvNeXt-Tiny outperforms all the other models in almost all natural image datasets (except CIFAR-10 and CIFAR-100). EfficientNetV2-S also performs well on atleast 5 out of 7 datasets. RegNetY-3.2GF also performs well on 5 of the 7 datasets. WaveMix performs the best in CIFAR-10 and CIFAR-100 but could not replicate the same performance on other datasets. 

ConvNeXt retains the good performance even amoung other domains such as textures, plant and astronomy. RegNet also performs well on texture, remote sensing and plant domains. We observe that SwinV2-Tiny performs the best in UC Merced Land Use dataset, WaveMix and ResNeXt top the EuroSAT dataset, DenseNet tops the PlantVillage dataset and WaveMix significantly outperforms all the other models in Galaxy10 DECals dataset. 

\subsection{Performance with less training data}

We observe from Table~\ref{tab:cifar10} that the models which are performing really well at 100\% of the training data are also performing better than other models at 10\% and 1\% of the training data. We can see from CIFAR-10 and Tiny Imaginary Dataset results that ConvNeXt, WaveMix, EfficientNet, and even RegNet are actually performing better than others even when the training data is significantly reduced. This shows that the fine-tuning superiority of these models is actually not dependent much on the amount of training data. Even with less data, they can generalize better and create better representations than other models.

\begin{figure}[]
\centering
\includegraphics[scale=0.7]{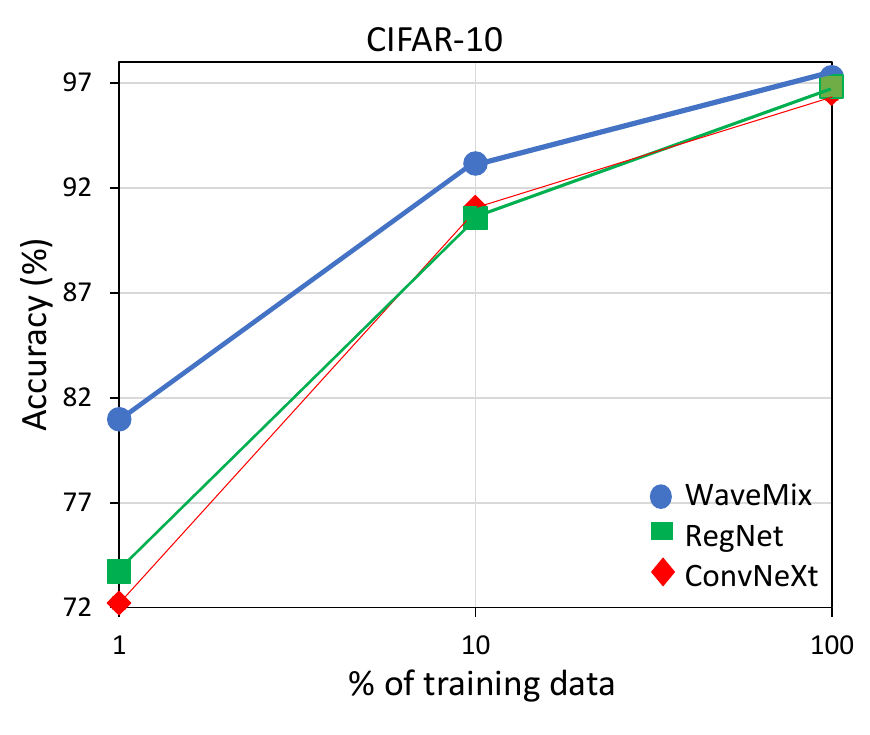}
\caption{Variation of accuracy of top 3 backbones with increasing training data. The model performance at 3 orders of training data, 1\% (500 images), 10\% (5000 images) and 100\% (50,000 images).
}
\label{fig:cifar10}
\end{figure}

\begin{figure*}[]
\centering
\includegraphics[scale=0.7]{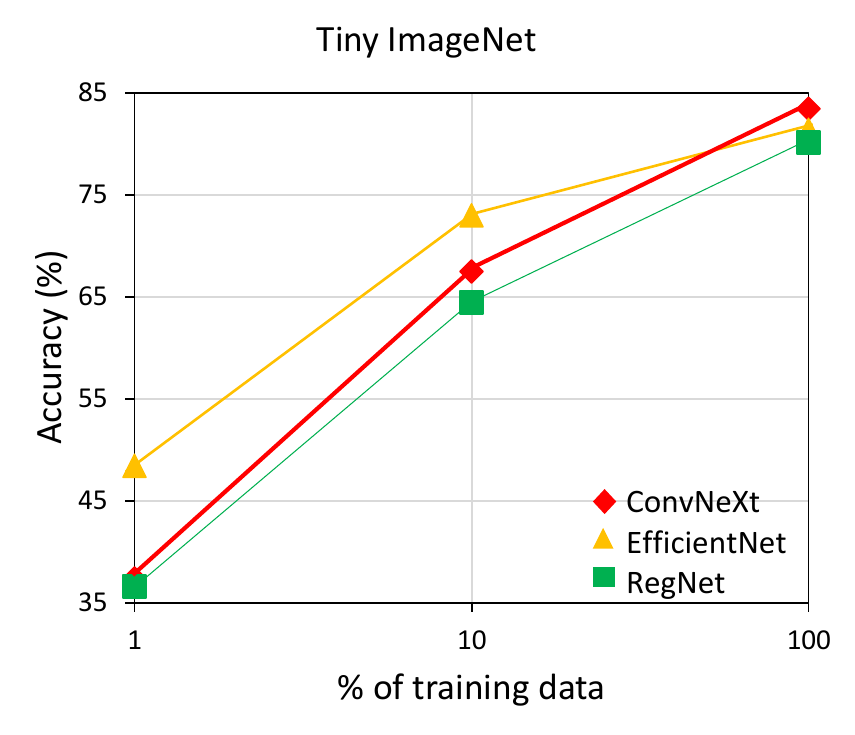}
\caption{Variation of accuracy of top 3 backbones with increasing training data. The model performance at 3 orders of training data, 1\% (1000 images), 10\% (10,000 images) and 100\% (100,000 images).
}
\label{fig:tiny}
\end{figure*}

Figure~\ref{fig:cifar10} and Figure~\ref{fig:tiny} shows the performance of top 3 models with training data. We see that the difference in accuracy amoung the three models is large and significant when the training data is less (1\%) and this difference reduces as the training data increases.  

We observe similar behaviour for other domains in Table~\ref{tab:other}. In plant datasets, the top performing models at 100\% of training data like ConvNeXt, EfficientNet, DenseNet, and RegNet are again performing better than the other models even at 10\% of the training data. Similar behaviour is also observed in radiology images, where ResNeXt and ResNeXt are performing better.

\begin{table}[]
\centering
\resizebox{\textwidth}{!}{%
\begin{tabular}{@{}lrrrrrrr@{}}
\toprule
\multicolumn{8}{c}{\textbf{Natural Images}} \\ \midrule
\textbf{Backbones} & \multicolumn{1}{l}{\textbf{Stanford Dogs}} & \multicolumn{1}{l}{\textbf{Flowers102}} & \multicolumn{1}{l}{\textbf{CUB200}} & \multicolumn{1}{l}{\textbf{Stanford Cars}} & \multicolumn{1}{l}{\textbf{Tiny ImageNet}} & \multicolumn{1}{l}{\textbf{CIFAR10}} & \multicolumn{1}{l}{\textbf{CIFAR100}} \\ \midrule
ResNet50 & 82.03 & 90.77 & 75.34 & 90.84 & 75.33 & 94.58 & 79.05 \\
WaveMix & 83.12 & 88.43 & 67.32 & 87.86 & 76.30 & \cellcolor[HTML]{34FF34}\textbf{97.26} & \cellcolor[HTML]{34FF34}\textbf{83.88} \\
ConvNeXt-Tiny & \cellcolor[HTML]{34FF34}\textbf{89.47} & \cellcolor[HTML]{34FF34}\textbf{94.41} & \cellcolor[HTML]{34FF34}\textbf{81.92} & \cellcolor[HTML]{34FF34}\textbf{92.26} & \cellcolor[HTML]{34FF34}\textbf{83.42} & \cellcolor[HTML]{D9EAD3}96.48 & 82.60 \\
Swin-Tiny & 83.81 & 90.07 & 78.16 & 90.04 & 76.85 & 95.38 & 81.25 \\
SwinV2-Tiny & 84.44 & 90.06 & 66.89 & \cellcolor[HTML]{D9EAD3}91.38 & 73.93 & 94.52 & 75.37 \\
EfficientNetV2-S & \cellcolor[HTML]{67FD9A}86.59 & \cellcolor[HTML]{67FD9A}93.65 & 79.10 & \cellcolor[HTML]{67FD9A}91.59 & \cellcolor[HTML]{67FD9A}81.35 & 96.42 & \cellcolor[HTML]{D9EAD3}83.20 \\
DenseNet-161 & 80.02 & 89.96 & 76.44 & 91.06 & 75.03 & 96.03 & 81.01 \\
MobileNetV3-large & 78.85 & 89.36 & 73.91 & 85.50 & 76.74 & 95.92 & 79.51 \\
RegNetY-3.2GF & \cellcolor[HTML]{D9EAD3}85.94 & \cellcolor[HTML]{D9EAD3}92.22 & \cellcolor[HTML]{67FD9A}81.24 &  91.33 & \cellcolor[HTML]{D9EAD3}80.14 & \cellcolor[HTML]{67FD9A}96.82 & 82.89 \\
ResNeXt-50 $32\times4$d & 85.09 & 91.78 & \cellcolor[HTML]{D9EAD3}78.29 & 90.22 & 77.86 & 96.16 & 81.30 \\
ShuffleNetV2 $2.0\times$ & 78.11 & 91.54 & 74.73 & 88.46 & 76.43 & 96.30 & \cellcolor[HTML]{67FD9A}83.32 \\ \bottomrule
\end{tabular}%
}
\vspace{1mm}
\caption{Top-1 classification accuracy for fine-tuning pre-trained backbones on natural image datasets. The top 3 results for each dataset are highlighted with best in green and bold, second best in lighter green, and third best in lightest green. }
\label{tab:natural}
\end{table}

\begin{table}[]
\centering
\begin{tabular}{@{}lrrrrrr@{}}
\toprule
\textbf{Domain} & \multicolumn{1}{c}{\textbf{Texture}} & \multicolumn{2}{c}{\textbf{Remote Sensing}} & \multicolumn{2}{c}{\textbf{Plant}} & \multicolumn{1}{c}{\textbf{Astronomy}} \\ \midrule
\textbf{Backbones} & \textbf{DTD} & \textbf{\begin{tabular}[c]{@{}r@{}}UC Merced \\ Land Use\end{tabular}} & \textbf{EuroSat} & \textbf{PlantVillage} & \textbf{PlantCLEF} & \textbf{Galaxy10} \\ \midrule
ResNet50 &  68.21 &  96.90 &  98.75 & \cellcolor[HTML]{67FD9A}99.83 &  80.60 &  84.80 \\
WaveMix &  68.25 &  97.72 & \cellcolor[HTML]{34FF34}\textbf{98.96} &  99.79 &  79.52 & \cellcolor[HTML]{34FF34}\textbf{95.10} \\
ConvNeXt-Tiny & \cellcolor[HTML]{34FF34}\textbf{73.70} & \cellcolor[HTML]{67FD9A}98.33 &  98.74 &  99.80 & \cellcolor[HTML]{67FD9A}82.71 & \cellcolor[HTML]{67FD9A}87.29 \\
Swin-Tiny &  70.15 &  97.86 &  98.52 &  99.70 &  79.03 &  84.64 \\
SwinV2-Tiny &  68.78 & \cellcolor[HTML]{34FF34}\textbf{98.81} &  98.50 &  99.76 &  78.16 &  83.15 \\
EfficientNetV2-S &  70.18 & \cellcolor[HTML]{D9EAD3}98.22 & \cellcolor[HTML]{67FD9A}98.88 & \cellcolor[HTML]{D9EAD3}99.81 &  81.09 &  84.75 \\
DenseNet-161 &  66.14 &  97.08 & \cellcolor[HTML]{D9EAD3}98.83 & \cellcolor[HTML]{34FF34}\textbf{99.88} &  76.49 & \cellcolor[HTML]{D9EAD3}86.70 \\
MobileNetV3-large &  68.69 &  97.14 &  98.72 &  99.80 &  75.52 &  82.47 \\
RegNetY-3.2GF & \cellcolor[HTML]{D9EAD3}71.16 & \cellcolor[HTML]{67FD9A}98.33 &  98.69 & \cellcolor[HTML]{67FD9A}99.83 & \cellcolor[HTML]{34FF34}\textbf{82.76} &  82.47 \\
ResNeXt-50 $32\times4$d & \cellcolor[HTML]{67FD9A}72.73 & \cellcolor[HTML]{67FD9A}98.33 & \cellcolor[HTML]{34FF34}\textbf{98.96} &  99.75 &  80.32 &  85.82 \\
ShuffleNetV2 $2.0\times$ &  68.64 &  97.86 &  98.65 &  99.70 & \cellcolor[HTML]{D9EAD3}81.76 &  83.69 \\ \bottomrule
\end{tabular}%
\vspace{1mm}
\caption{Top-1 classification accuracy for fine-tuning pre-trained backbones on different domain datasets. The top 3 results are highlighted with best in green and bold, light green, and third best in lightest green.  }
\label{tab:domain}
\end{table}

\begin{table}[]
\centering
\begin{tabular}{@{}lrrrrr@{}}
\toprule
\multicolumn{6}{c}{\textbf{Medical Images (Histopathology) BreakHis Dataset}} \\ \midrule
\textbf{Backbones} & \textbf{$40\times$} & \textbf{$100\times$} & \textbf{$200\times$} & \textbf{$400\times$} & \textbf{Average} \\ \midrule
ResNet50 &  97.91 & \cellcolor[HTML]{34FF34}\textbf{99.53} &  99.22 &  98.44 &  98.78 \\
WaveMix &  99.42 & \cellcolor[HTML]{67FD9A}99.42 & \cellcolor[HTML]{D9EAD3}99.38 & \cellcolor[HTML]{34FF34}\textbf{99.35} & \cellcolor[HTML]{34FF34}\textbf{99.39} \\
ConvNeXt-Tiny &  95.10 &  90.73 &  88.59 &  88.72 &  90.79 \\
Swin-Tiny &  88.11 &  93.89 &  90.00 &  82.61 &  88.65 \\
SwinV2-Tiny &  89.96 &  92.44 &  88.65 &  83.83 &  88.72 \\
EfficientNetV2-S &  99.44 & \cellcolor[HTML]{67FD9A}99.42 &  99.11 & \cellcolor[HTML]{67FD9A}99.06 & \cellcolor[HTML]{67FD9A}99.26 \\
DenseNet-161 &  98.62 & \cellcolor[HTML]{D9EAD3}99.24 &  99.28 &  98.20 &  98.84 \\
MobileNetV3-Large & \cellcolor[HTML]{67FD9A}99.56 &  98.24 & \cellcolor[HTML]{34FF34}\textbf{99.50} & \cellcolor[HTML]{D9EAD3}98.94 &  99.06 \\
RegNetY-3.2GF & \cellcolor[HTML]{34FF34}\textbf{99.84} &  99.22 & \cellcolor[HTML]{67FD9A}99.48 &  98.02 & \cellcolor[HTML]{D9EAD3}99.14 \\
ResNeXt-50 $32\times4$d & \cellcolor[HTML]{D9EAD3}99.46 &  99.22 &  99.34 &  98.05 &  99.02 \\
ShuffleNetV2 $2.0\times$ &  99.38 &  98.59 &  99.22 &  98.02 &  98.80 \\ \bottomrule
\end{tabular}
\vspace{1mm}
\caption{Top-1 classification accuracy for fine-tuning pre-trained backbones on medical datasets. The top 3 results are highlighted with best in green and bold, light green, and third best in lightest green.  }
\label{tab: medical}
\end{table}

\begin{table}[]
\centering
\begin{tabular}{@{}lrrrrrr@{}}
\toprule
\multicolumn{1}{c}{} & \multicolumn{3}{c}{CIFAR-10} & \multicolumn{3}{c}{Tiny ImageNet} \\
Number of training images & 50,000 & 5,000 & 500 & 100,000 & 10,000 & 1,000 \\
Percentage of training data & 100\% & 10\% & 1\% & 100\% & 10\% & 1\% \\ \midrule
ResNet & 94.58 &  86.5 & 65.42 & 76.30 & 66.56 & 28.11 \\
WaveMix & \cellcolor[HTML]{ 34FF34}\textbf{97.26} & \cellcolor[HTML]{ 34FF34}\textbf{93.16} & \cellcolor[HTML]{ 34FF34}\textbf{80.98} & 75.33 &  55.15 & 25.24 \\
ConvNeXt & \cellcolor[HTML]{D9EAD3}96.48 & \cellcolor[HTML]{67FD9A}91.06 & 72.22 & \cellcolor[HTML]{ 34FF34}\textbf{83.42} & \cellcolor[HTML]{67FD9A}67.47 & \cellcolor[HTML]{67FD9A}37.38 \\
Swin & 95.38 &  85.64 & 39.77 & 76.85 &  48.50 & 25.62 \\
SwinV2 & 94.52 & 85.66 & 46.51 & 73.93 &  39.31 & 12.29 \\
EfficientNet & 96.42 &  89.44 & \cellcolor[HTML]{D9EAD3}77.06 & \cellcolor[HTML]{67FD9A}81.35 & \cellcolor[HTML]{ 34FF34}\textbf{72.97} & \cellcolor[HTML]{ 34FF34}\textbf{48.42} \\
DenseNet & 96.03 &  86.92 & 64.83 & 75.03 &  51.89 & 18.00 \\
MobileNet & 95.92 &  88.85 & 61.27 & 76.74 &  62.70 & 31.08 \\
RegNet & \cellcolor[HTML]{67FD9A}96.82 & \cellcolor[HTML]{D9EAD3}90.55 & 73.71 & \cellcolor[HTML]{D9EAD3}80.14 &  64.40 & \cellcolor[HTML]{D9EAD3}36.56 \\
ResNeXt & 96.16 &  90.24 & 73.12 & 77.86 & \cellcolor[HTML]{D9EAD3}67.01 & 34.40 \\
ShuffleNet & 96.30 & 90.62 & \cellcolor[HTML]{67FD9A}77.45 & 76.43 &  61.65 & 29.89 \\ \bottomrule
\end{tabular}%
\vspace{1mm}
\caption{The variation of accuracy of backbones with reduction of training data. Performance at 100\%, 10\% and 1\% of training data for CIFAR-10 and Tiny ImageNet datasets }
\label{tab:cifar10}
\end{table}

\begin{table}[]
\centering
\resizebox{\textwidth}{!}{%
\begin{tabular}{@{}lrrrrrrrrrr@{}}
\toprule
\multicolumn{1}{c}{} & \multicolumn{2}{c}{PlantVillage} & \multicolumn{2}{c}{PlantCLEF} & \multicolumn{2}{c}{RSNA} & \multicolumn{2}{c}{Food-101} & \multicolumn{2}{c}{EuroSAT} \\
Number of training images & 44,343 & 4,434 & 10,455 & 1,045 & 24,181 & 2,418 & 75,750 & 7,575 & 18.900 & 1,890 \\
Percentage of training data & 100\% & 10\% & 100\% & 10\% & 100\% & 10\% & 100\% & 10\% & 100\% & 10\% \\ \midrule
ResNet & \cellcolor[HTML]{67FD9A}99.83 & 98.07 & 80.60 & 57.47 & \cellcolor[HTML]{D9EAD3}87.3 & \cellcolor[HTML]{D9EAD3}81.26 & 88.36 & 74.26 & 98.75 & 96.49 \\
WaveMix & 99.79 & 97.39 & 79.52 & 57.64 & 86.94 & 80.93 & 85.88 & 68.86 & 98.96 & 96.48 \\
ConvNeXt & 99.80 & \cellcolor[HTML]{ 34FF34}\textbf{98.92} & \cellcolor[HTML]{67FD9A}82.71 & \cellcolor[HTML]{ 34FF34}\textbf{64.68} & 86.71 & 80.03 & \cellcolor[HTML]{D9EAD3}89.43 & \cellcolor[HTML]{ 34FF34}\textbf{78.47} & 98.74 & \cellcolor[HTML]{ 34FF34}\textbf{97.29} \\
Swin & 99.70 & 97.75 & 79.03 & 48.57 & 70.36 & 68.22 & 88.81 & 76.21 & 98.52 & 95.95 \\
SwinV2 & 99.76 & 97.72 & 78.16 & 21.47 & 86.38 & 67.63 & 89.04 & 74.17 & 98.50 & 96.48 \\
EfficientNet & \cellcolor[HTML]{D9EAD3}99.81 & \cellcolor[HTML]{D9EAD3}98.42 & 81.09 & \cellcolor[HTML]{D9EAD3}64.4 & 87.06 & 79.96 & \cellcolor[HTML]{67FD9A}89.84 & \cellcolor[HTML]{D9EAD3}76.68 & \cellcolor[HTML]{67FD9A}98.88 & \cellcolor[HTML]{67FD9A}97.23 \\
DenseNet & \cellcolor[HTML]{ 34FF34}\textbf{99.88} & \cellcolor[HTML]{67FD9A}98.57 & 76.49 & 58.59 & 87.1 & \cellcolor[HTML]{ 34FF34}\textbf{81.51} & 87.02 & 72.44 & \cellcolor[HTML]{D9EAD3}98.83 & \cellcolor[HTML]{D9EAD3}96.86 \\
MobileNet & 99.80 & 98.29 & 75.52 & 54.93 & 87.05 & 79.55 & 85.73 & 71.12 & 98.72 & 96.35 \\
RegNet & \cellcolor[HTML]{67FD9A}99.83 & 98.25 & \cellcolor[HTML]{ 34FF34}\textbf{82.76} & \cellcolor[HTML]{67FD9A}64.53 & 87.05 & 79.56 & \cellcolor[HTML]{ 34FF34}\textbf{90.4} & \cellcolor[HTML]{67FD9A}77.89 & 98.69 & 96.49 \\
ResNeXt & 99.75 & 98.01 & 80.32 & 63.61 & \cellcolor[HTML]{ 34FF34}\textbf{87.89} & \cellcolor[HTML]{67FD9A}81.42 & 88.89 & 74.69 & \cellcolor[HTML]{ 34FF34}\textbf{98.96} & 96.53 \\
ShuffleNet & 99.70 & 98.30 & \cellcolor[HTML]{D9EAD3}81.76 & 58.69 & \cellcolor[HTML]{67FD9A}87.53 & 80.42 & 86.21 & 70.97 & 98.65 & 96.00 \\ \bottomrule
\end{tabular}%
}
\vspace{1mm}
\caption{The variation of accuracy of backbones with reduction of training data. Performance at 100\% and 10\% training data.}
\label{tab:other}
\end{table}

\section{Observations}

$\bigtriangleup$ \textbf{Higher pre-training accuracy in ImageNet does not mean higher fine-tuning accuracy.} From Table~\ref{tab:imagenet} we can see that amoung all the backbones we used for our experiments, EfficientNetV2-S has the highest pre-training ImageNet-1k accuracy (84.23\%). Even though the fine-tuning performance of EfficientNet was decent, it did not perform the best in any of the domains or datasets we evaluated on. Even amoung datasets such as Tiny ImageNet and Stanford Dogs whose images are sourced from ImageNet dataset, we find that ConvNeXt outperforms EfficientNet. Therefore, we recommend the practitioners to not use the pre-training accuracy as a metric to choose the backbone. 

$\bigtriangleup$ \textbf{Convolutional models strongly outperform transformers in resource-efficient low-data fine-tuning.} Even though Swin transformer has a hierarchical structure capable of exploiting the spacial inductive bias~\cite{goldblum2023battle}, they still perform poorly in almost all our tasks compared to modern CNN architectures such as ConvNeXt (which was designed based on Swin transformer). Hence, we recommend that for the low data fine-tune regime, it is better to avoid using transformer architectures like Swin and use pure CNN backbones such as ConvNeXt, EfficientNet or RegNet. 

$\bigtriangleup$ \textbf{ConvNeXt architecture consistently outperforms other models when fine-tuning on natural image datasets.}   This superior performance can be attributed to the innovative design of ConvNeXt, which integrates architectural advancements that bridge the gap between traditional convolutional networks, like ResNet, and modern Swin transformer models. ConvNeXt retains the beneficial convolutional inductive bias, enabling it to learn more effectively than other convolution-based architectures and even some attention-based transformer models. Our data clearly indicates that for natural images, ConvNeXt stands out as the best model, delivering exceptional fine-tuning performance.

$\bigtriangleup$ \textbf{RegNet and EfficientNet models are excellent choices for fine-tuning across a wide range of image domains.} While ConvNeXt excels predominantly with natural images, EfficientNet closely follows in performance, and RegNet also shows strong results in this domain. However, the versatility of RegNet and EfficientNet extends beyond natural images. Our experiments reveal that these models also perform exceptionally well on diverse datasets, including remote sensing images, plant datasets, and medical images, such as histopathology images. Therefore, we recommend practitioners to consider RegNet and EfficientNet when working with datasets beyond natural images, as their adaptability and robust performance across various domains make them valuable tools for fine-tuning tasks.

$\bigtriangleup$ \textbf{ShuffleNet is a better choice than MobileNet when very light-weight models are needed.} When we observe the performance of very lightweight models, specifically those with a model size of less than 50 MB, which are ideal for on-device applications, we find that ShuffleNetV2 generally outperforms MobileNetV3 across multiple domains. Although MobileNetV3 has shown better performance on medical domain, ShuffleNetV2 demonstrates more consistent and slightly superior performance across a broader range of image domains. Therefore, we recommend ShuffleNet v2 as a better choice for practitioners dealing with on-device applications, where fine-tuning a model on a domain-specific dataset is required.

$\bigtriangleup$ \textbf{WaveMix performs well in datasets where multi-resolution token-mixing aid in learning} WaveMix outperforms all other models (9\% increase from the second best, ConvNeXt) in galaxy morphology classification. WaveMix also performs well in medical domain performing better than all other datasets, and also maintaining the performance across different magnification. WaveMix, which uses 2D-DWT might possess inductive bias that can analyse the domains of astronomy and medical images better than other convolutional models due its multi-resolution token-mixing. We recommend using WaveMx in domains where features across different resolutions are needed for better performance. WaveMix also performs better than ConvNeXt in CIFAR-10 and CIFAR-100 datasets, which were actually low resolution natural images ($32\times32$) which were resized to ($256\times256$) for our training. Similarly, it gives best performance in remote sensing dataset, EuroSAT, whose images ($64\times64$) were resized to ($256\times256$) for our training. WaveMix is also state-of-the-art in many low-resolution datasets such as EMNIST ($28\times28$). The only low resolution image dataset where WaveMix did not perform well is Tiny ImageNet whose images ($64\times64$) were also resized to ($256\times256$). We attribute this to the fact that Tiny ImageNet is an ImageNet-1k subset and the other models which has better performance in ImageNet-1k naturally performed better. So, we recommend using WaveMix for low-resolution image datasets. 

$\bigtriangleup$ \textbf{Age of ResNet dominance is over.} Our results also reveal that ResNet is no longer competitive compared to these modern architectures in any domain. Our experimental data shows that for natural images, ResNet does not even rank among the top three performers. It significantly lags behind newer models such as ConvNeXt. While ResNet does perform relatively well on one medical dataset, even in these cases, other models achieve similar performance levels. Therefore, it is clear that practitioners should transition from using ResNet to these newer architectures to achieve better results in their fine-tuning tasks.

$\bigtriangleup$ \textbf{ConvNeXt and Swin transformers perform poorly in medical domain.} Our results show that both the architectures perform perform really bad compared to all the other models when used in medical domain. Therefore, we suggest practitioners working in medical domain to not use these models despite their high performance in other domains.

$\bigtriangleup$ \textbf{Most of the top models in every domain retain their higher performance even with less training data.} We find that even when we fine-tune with a small percentage of training data (even 1\% $\sim1000$ images), the models which performed well with full training set still retained their superiority. This points to the presence of a domain specific inductive bias present in these models since they can learn better representations with very less data. The failure of these models to perform well on other domains similarly with less training data also alludes to this inductive bias. 

\begin{table}[]
\centering
\resizebox{\textwidth}{!}{%
\begin{tabular}{@{}llrrllrrlll@{}}
\toprule
Backbone & Block & \begin{tabular}[c]{@{}r@{}}Kernels \\ in block\end{tabular} & \begin{tabular}[c]{@{}r@{}}\# \\ Stages\end{tabular} & \#Blocks/stage & Non-linearity & CEF & \multicolumn{1}{c}{Stem} & \multicolumn{1}{c}{DConv} & \multicolumn{1}{c}{SE} & Norm \\ \midrule
ResNet-50 & Bottleneck & 1$\times$1, 3$\times$3, 1$\times$1 & 4 & 3,4,6,3 & ReLU & 4 & 7$\times$7, /2 & No & No & BN \\
DenseNet-161 & Inverted BN & 1$\times$1, 3$\times$3 & 4 & 6,12,36,24 & ReLU & 4 & 7$\times$7, /2 & No & No & BN \\
ResNeXt-50-32$\times$4d & Bottleneck & 1$\times$1, 3$\times$3, 1$\times$1 & 4 & 3,4,6,3 & ReLU & 2 & 7$\times$7, /2 & No & No & BN \\
ShuffleNetV2-$\times$2.0 & Isotropic & 1$\times$1, 3$\times$3, 1$\times$1 & 3 & 4,8,4 & ReLU & 1 & 3$\times$3, /2 & Yes & No & BN \\
MobileNetV3-Large & Inverted BN & 1$\times$1, 3$\times$3/5$\times$5, 1$\times$1 & 4 & 2,3,6,3 & ReLU, SiLU & 3-6 & 3$\times$3, /2 & Yes & Yes & BN \\
RegNet-Y3-2GF & Isotropic & 1$\times$1, 3$\times$3, 1$\times$1 & 4 & 2, 5, 13, 1 & ReLU & 1 & 3$\times$3, /2 & No & Yes & BN \\
EfficientNet-V2-S & Inverted BN & 1$\times$1, 3$\times$3, 1$\times$1 & 6 & 2, 4, 4, 6, 9, 15 & SiLU & 4-6 & 3$\times$3, /2 & Yes & Yes & BN \\
WaveMix-192/16 & Tokenmixer & 1$\times$1, 1$\times$1, 1$\times$1 & 1 & 16 & GELU & 2 & 4$\times$4, /4 & No & No & BN \\
ConvNeXt-Tiny & Tokenmixer & 7$\times$7, 1$\times$1, 1$\times$1 & 4 & 3,3,9,3 & GELU & 4 & 4$\times$4, /4 & Yes & No & LN \\ \bottomrule
\vspace{1mm}
\end{tabular}%
}
\caption{Architectural comparison of all the convolutional backbones showing the block structure of the residual block (eg. BottleNeck (BN)), kernels used in each block in sequence, number of stages in pyramidal models, number of blocks in each stage in pyramidal models, the non-linearities used in networks blocks, the channel expansion factor (CEF) in each block, the kernel and stride used in the initial stem layer, whether depth-wise convolutions (DConv) and squeeze-excitation (SE) operations were used in blocks, and the normalization used in blocks (whether BatchNorm (BN) or LayerNorm (LN)). The architectures are ordered in descending order of age.}
\label{tab:arch}
\end{table}

\section{Architectural Discussions}

Table~\ref{tab:arch} shows the architectural details of all the convolutional architectures used in our experiments from oldest to latest. From the Table, we can see certain architectural trends that has improved the performance of the newer models compared to older ones. All top performing models in all the domains are the latest models such as RegNet, EfficientNet, WaveMix and ConvNeXt. These observations can be used to develop better resource-efficient backbones for fine-tuning tasks. 

Most of the models use a $1\times1$, $3\times3$, $1\times1$ kernel structure in their blocks with a normalisation and activation following each convolutional operation. MobileNet uses $5\times5$ kernels towards the final stages, while ConvNeXt uses $7\times7$ kernel for spatial token-mixing in each block. All architectures use $1\times1$ convolution to change the number of channels. 

While increasing the channel dimensions inside bottlenecks/inverted bottlenecks, almost all models use a channel expansion/contraction factor of 4. MobileNet and EfficientNet uses even higher values up to 6. 

All models use adaptive average pooling in the classification head to down-sample the output feature maps from the final stage to $1\times1$ spatial resolution.

$\bigtriangleup$ \textbf{Inverted bottleneck design is better.} The older models used bottleneck blocks where channel dimension was reduced using $1\times1$ convolutions before being passed to the main $3\times3$ convolutions to increase efficiency. Newer architectures such as EfficientNet and MobileNet use an inverted bottleneck structure where the $1\times1$ convolutions increases the channel dimension before $3\times3$ convolutions. The increase in parameters is offset mainly by using the parameter efficient depth-wise convolution which also increases performance. Models like ShuffleNet and RegNet keeps the channel dimension invariant throughout the block resulting in an isotropic design. WaveMix and ConvNeXt follow the token-mixing structure where spatial and channel mixing is performed separately, with spacial mixing done with token-mixing operations such as wavelet transform and large kernel depth-wise convolution respectively. The channel dimension is expanded only for channel mixing using $1\times1$ convolutions in an inverted bottleneck design.

$\bigtriangleup$ \textbf{Depth-wise convolution replacing regular convolution.} Depth-wise convolutions have fewer parameters and requires less operations than regular convolutions. Most of the newer models such as ShuffleNet, MobileNet and EfficientNet replace the $3\times3$ kernel regular convolution with $3\times3$ kernel depth-wise convolution. ConvNeXt uses $7\times7$ kernel depth-wise convolution for efficient spacial token-mixing. All $1\times1$ convolutions in all these models are point-wise convolutions (regular convolutions). SiLU activation has been observed to give better performance when used with depth-wise convolutions~\cite{radosavovic2020designing}. 

$\bigtriangleup$ \textbf{4 stage models with largest number of blocks in penultimate stage.} From old to new models, 4 stage structure has been most popular and has resulted in good performance. Other than EfficientNet with 6 stages, ShuffleNet with 3 stages and WaveMix with a single stage, all other backbone has 4 stages. The design principle of increasing the number of blocks with each subsequent stage with the exception of final stage is also observed in most models. Only Efficient has more blocks in the final stage than the penultimate stage. Most of the 4 stage models use a channel expansion factor greater than or equal to 2 during feature resolution down-sampling between stages to increase the number of channels.

$\bigtriangleup$ \textbf{Strided convolution for down-sampling.} With the exception of DenseNet which used average pooling for down-sampling the feature resolution between different stages of the network, all other convolutional models used strided convolutions with stride 2 for feature resolution down-sampling. So strided convolutions are better for model performance than pooling operations. 

$\bigtriangleup$ \textbf{Small kernel stem layer.} Stem layer is the initial layer used to down-sample the full size input image whose output is then sent to the body of architectures which has multiple stages. Compared to older models which used large $7\times7$ kernel with stride 2, newer models use $3\times3$ and $4\times4$ kernels. The token-mixers use patchify stem ($4\times4$ non-overlapping convolutions). 

$\bigtriangleup$ \textbf{SiLU and GELU replacing ReLU.} Older models used ReLU activation after convolution operations. The newer models use a combination of ReLU and SiLU activations while the token-mixers inspired from transformer architecture uses GELU activation. Compared to other convolutional models which uses 2 to 3 activations per block, token-mixers use much less activations (one per block). 

$\bigtriangleup$ \textbf{Addition of squeeze-excitation (SE) module.~\cite{hu2019squeezeandexcitationnetworks}} SE module enhance CNNs by focusing on the most informative features through explicit modeling of channel interdependencies and can be easily integrated into various architectures without significant changes. The top performing architectures such as RegNet and EfficientNet used SE modules to improve performance.

\section{Conclusions}
\label{sec:conclusions}

Computer vision backbones are critical for fine-tuning on domain specific datasets, particularly when using pre-trained backbones available in libraries such as Torchvision. In our study, we have compared the performance of various lightweight, resource-efficient backbones across different domains, including medical images, natural images, astronomy images, plant images, and remote sensing images.

Our analysis revealed that modern architectural models like ConvNeXt, EfficientNet, and RegNet excel in handling multiple domains images. Additionally, models with specific inductive biases, such as WaveMix, are particularly useful for tasks requiring multi-resolution analysis. Among lightweight, on-device models, ShuffleNet consistently outperformed MobileNet in fine-tuning tasks. We also observed that transformer-based or attention-based models, such as Swin Transformer, do not perform well when the fine-tuning dataset is small.

Based on our findings, we offer practical recommendations for using pre-trained computer vision backbones. Our comparative analysis on image classification fine-tuning serves as a valuable guide for practitioners and researchers alike, who aim to optimize their models for various image domains. We hope that our work will contribute to the development of better model architectures capable of performing well across diverse image datasets.

\textbf{Limitations:} We restricted our comparison to models available in Torchvision, focusing specifically on lightweight and resource-efficient architectures. Consequently, we did not analyze any larger models with more than 30 million parameters, limiting our ability to test the scalability of these models with larger fine-tuning datasets. Furthermore, our analysis was confined to fine-tuning datasets containing fewer than 100,000 images, which may not fully represent scenarios involving significantly larger datasets and scalability of these backbones. 

Another limitation of our work is that we exclusively focused on the computational task of image classification. We did not extend our analysis to other important tasks in computer vision, such as object detection or image retrieval. The performance of various backbones on these other tasks remains unexplored in our study. While we hope that there might be some correlation with the performance we observed to these other computer vision tasks, this remains speculative and requires further investigation to confirm.

\textbf{Computation Cost and Carbon Footprint:} The experiments in this paper took a cumulative 1500 GPU hours on NVIDIA RTX A100 cards. Assuming the GPUs were running with an average carbon efficiency of 0.37 kgCO$_2$eq/kWh, the total emissions are estimated to be 222 kgCO$_2$eq.

\bibliographystyle{unsrt}  
\bibliography{references}  

\appendix

\end{document}